\documentclass[runningheads]{llncs}
\usepackage{graphicx}
\usepackage{comment}
\usepackage{amsmath,amssymb} 
\usepackage{color}
\usepackage{subfigure}
\usepackage{booktabs}

\begin{document}
\pagestyle{headings}
\mainmatter
\def\ECCVSubNumber{690}  

\title{Learning a Domain Classifier Bank for Unsupervised Adaptive Object Detection}

\titlerunning{Domain Classifier Bank for Unsupervised Adaptive Object Detection}
%
\author{
Sanli Tang\inst{1} \and
Zhanzhan Cheng\inst{21} \and
Shiliang Pu\inst{1} \and \\
Dashan Guo\inst{1} \and
Yi Niu\inst{1} \and
Fei Wu\inst{2}
}
%
%
\institute{
Hikvision Research Institute, China
\and
Zhejiang University, Hangzhou, China\\
}

\authorrunning{Sanli Tang, Zhanzhan Cheng et al.}
\maketitle

\begin{abstract}
In real applications, object detectors based on deep networks still face challenges of the large domain gap between the labeled training data and unlabeled testing data.
To reduce the gap, recent techniques are proposed by aligning the image/instance-level features between source and unlabeled target domains.
However, these methods suffer from the suboptimal problem mainly because of {ignoring the category information of object instances.}
To tackle this issue, we develop a fine-grained domain alignment approach with a well-designed domain classifier bank that achieves the instance-level alignment respecting to their categories.
Specifically, we first employ the mean teacher paradigm to generate pseudo labels for unlabeled samples.
Then we implement the class-level domain classifiers and group them together, called domain classifier bank, in which each domain classifier is responsible for aligning features of a specific class. 
We assemble the bare object detector with the proposed fine-grained domain alignment mechanism as the adaptive detector, and optimize it with a developed crossed adaptive weighting mechanism.
Extensive experiments on three popular transferring benchmarks demonstrate the effectiveness of our method and achieve the new remarkable state-of-the-arts. 

\keywords{Object Detection, Domain Classifier Bank, Domain Alignment}
\end{abstract}

\section{Introduction}
Deep neural networks have shown great power on various tasks \cite{Fast,ImageNet,Segmentation}, but heavily rely on the amount of labelled data.
In the real world, it is much costly to annotate a large-scale dataset especially for object detection tasks.
Thus, training a model on label-rich dataset (source domain) and then transferring to the unlabelled data (target domain), namely the unsupervised domain adaptation (\emph{abbr.} UDA), is a promising solution \cite{Adapting,Subspace}.
For example, the auto-annotated vehicles from self-driving simulation system such as GTA can be used to help improve the vehicle detection performance in the real-world.

Early researches aim at shrinking the domain gap \cite{DomainGAP} by aligning the model's activating responses to data from both the source and the target domains \cite{SEDA,Subspace,MDI}.
Inspired by the adversarial training techniques \cite{DAT,ADDA} in image classification task, recent two methods \cite{DA,SWDA} attempted to directly incorporate the bare detectors with domain classifiers to extract the image-level or instance-level domain-invariant features, and achieved significant results.
\begin{figure}[t]
\centering
\subfigure[]{
\begin{minipage}{0.35\textwidth}
\centering
\includegraphics[height=2.9cm]{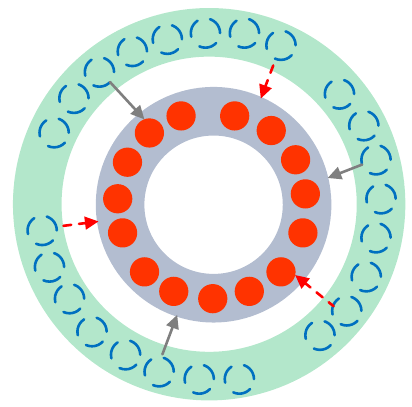}
\end{minipage}
}%
\subfigure[]{
\begin{minipage}{0.35\textwidth}
\centering
\includegraphics[height=2.9cm]{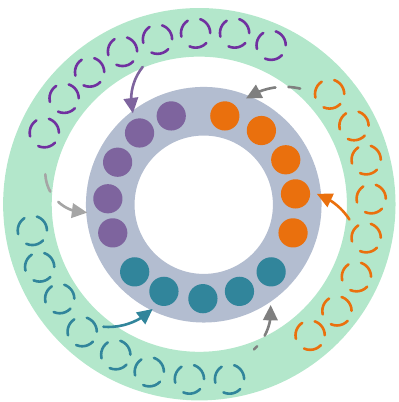}
\end{minipage}
}%
\subfigure{
\begin{minipage}{0.2\textwidth}
\centering
\includegraphics[height=2.9cm]{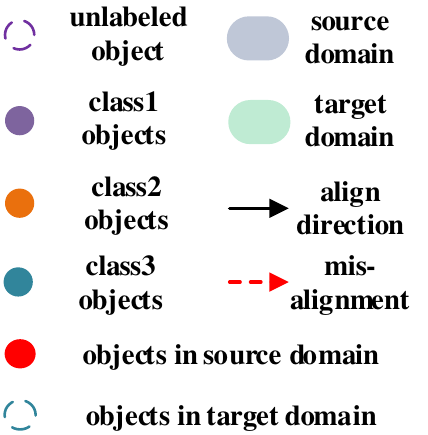}
\end{minipage}
}
\caption{Illustration of domain alignment. (a) shows the instance-level feature alignment regarding to the object instances in source or target domain where all instances share the same class.
(b) shows the class-level alignment by considering the category labels of object instances.
The hollow and solid circles are corresponding to the instance labels in target and source domains, respectively.
The arrows represent the aligning directions.}
\label{FigExample}
\end{figure}

However, the image-level domain alignment (\emph{abbr.} ImDA) strategy, such as \cite{SWDA}, takes no account of the significant difference in object number, size and even the layouts in different domains.
This method can be treated only as a coarse solution and has very limited effects.
Though the instance-level (\emph{abbr.} InDA) strategy, such as \cite{DA}, considers the issues in ImDA, it still suffers from the suboptimal problem due to lacking of considering their category information.
To be specific, it is unreasonable to train domain classifiers by regarding all instances from source/target domain as the same class, which prevents the model from drawing a clear distinction among different object categories. As a result, the detector is easy to be confused.
Figure \ref{FigExample} (a) illustrates InDA that each instance from target domain is aligned to its closest instance from source domain, which inevitably leads to the misalignment between different instance categories.

Considering above issues, a more promising way is to align instance features according to their ground truth or pseudo labels. 
It means the detected objects with {higher confidence} in target domain should be paid more attention on aligning the instance features regarding to their categories. 
For example in Figure \ref{FigExample}(b), the target objects predicted as class-2 (denoted as brown hollow circles) should be aligned to the class-2 objects in source domain (denoted as brown solid circles).
Here, we treat it as class-level domain alignment.
To achieve class-level alignment in all classes, a group of domain classifiers can be established as a domain classifier bank in which each classifier takes the charge of aligning features of a specific class.
In this way, the closest pairs of instances between the source and target domains refer to those objects sharing the same class from the perspective of the detector.
Note that, predicted results of domain classifiers also reveal the effects of feature alignment, i.e., the more extent of alignment, the more confused prediction will be made by the domain classifiers, as addressed in \cite{TADA}. Thereby, for those well-aligned features, their pseudo labels could be added more weight when training the detector on unlabeled data from the target domain.
Then the detection performance can be further enhanced. 

In this paper, we propose a fine-grained unsupervised domain adaptation method for object detection, which consists of a domain classifier bank integrating with a teacher-student framework.
Concretely, we group the class-level domain classifiers together to form as a bank, named as DCBank, in which each classifier is responsible for aligning features of the specific class between the source and target domains.
Since images in target domain are unlabeled, mean teacher \cite{MT} is employed to provide pseudo labels, e.g. the locations and classes of the objects.
{The generated pseudo labels can be used to train the DCBank. 
}
We integrate the bare object detector with mean teacher as well as the DCBank into an unsupervised adaptive detection framework named as MDBank, and optimize it with a crossed adaptive weighting mechanism.
Here, the crossed adaptive weights are calculated from the predicting confidence of the detector as well as the entropy of the DCBank, and can improve both the detector and the DCBank.


The contributions are summarized as follows:
(1) We address the class-level domain adaptation problem, and develop the domain classifier bank mechanism to align instance-level features according to their categories.
(2) We assemble a bare detector with mean teacher as well as the designed DCBank into an adaptive detection framework. The whole framework is jointly optimized with a crossed weighting strategy which can improve both the detector and DCBank.
(3) Extensive experiments on three popular datasets demonstrate the effectiveness of our method. 

\section{Related works}
\subsection{Object Detection}
Object detectors based on deep neural networks can be roughly divided into two categories: the two-stage and the one-stage.
Faster R-CNN \cite{Faster} is a representative two-stage detector, where a Region Proposal Network (RPN) is designed to provide object proposals, e.g. the coarse bounding boxes and the probabilities of their being the foreground category. Then the cropped and resized features are fed into a classifier in the second stage to predict their categories and refined locations.
A series of improvements \cite{CascadeRCNN,MaskRCNN,LightHeadRCNN} based on Faster R-CNN have also been explored to further boost the performance.
While for one-stage detectors, YOLO \cite{Yolo} directly regressed the bounding boxes and the confidence of being multiple categories, which achieved competitive performance in a high efficiency manner. SSD \cite{SSD} aimed to increase the detection rates of objects in different scales, especially the small ones by predicting from multiple feature maps at different resolutions. After that, \cite{RetinaNet,Yolov3,FCOS} further advanced the one-stage detectors by revising the network structure or applying delicate training skills.


\subsection{Domain Adaptation}
Many researches \cite{SEDA,ADGAN,MDI} on domain adaptation struggle for bridging the gap between the source and target domain. Earlier works tried to minimize the discrepancy between two domains, which was defined in statistics, e.g. the Maximum Mean Discrepancy (MMD) \cite{MMD,DTL,WMMD} or CORAL distance \cite{Coral,DeepCoral}. Recent methods \cite{ADDA} based on adversarial training aligned the feature distribution by cheating domain classifiers that were trained to distinguish the image features from different domain. \cite{TADA} proposed the transferable attention that assigns different weights to feature maps according to the predicting confidence of the domain classifiers.
Derived from Mean Teacher \cite{MT} in semi-supervised learning, self-ensembling \cite{SEDA} was proposed to extract domain-invariant features by minimizing the outputs of the teacher and the student with augmented inputs. All above methods were examined in image classification tasks.

Recently, researchers start to pay attention to domain adaptation in object detection.
In general, existing methods could be summarized as three types of domain alignments: the input-level, the feature-level and the output-level.
(1) The input-level aligning techniques usually adopted generative models to directly transfer input images from the source domain to the target domain while keeping the labels unchanged {\cite{PDA}}. Then, the generated labeled images could be utilized to train a detector in a fully supervised manner.
(2) For feature-level alignment, {\cite{DA}} aligned both top features in backbone and instance features by adversarial training, where the instance-level domain classifier treated the object features as the same class only if they come from the same domain.
Strong weak domain adaptation (SWDA) \cite{SWDA} argued that the precisely matching on global features was likely to hurt the performance confronting with large domain gap, and adopted a weak image-level domain classifier by focal loss \cite{RetinaNet} to align features. 
(3) For output alignment, mean teacher with object relation (MTOR) \cite{MTOR} made three kinds of consistency regularization based on two relational graphs in teacher and student networks, which showed a promising way of self-ensembling framework in UDA object detection tasks.
\cite{RobustFaster} addressed the UDA detection problem by training a detector on the target domain with noisy object bounding box. 
Here, we focus on the feature-level domain alignment.

Unlike previous feature-level alignment methods in which aligning domain features regardless of their classes, in this paper, we try to achieve a fine-grained instance-level domain alignment by regarding to instance  categories.

\section{Preliminary Work}
\begin{figure}[t]
\centering
\includegraphics[width=8cm]{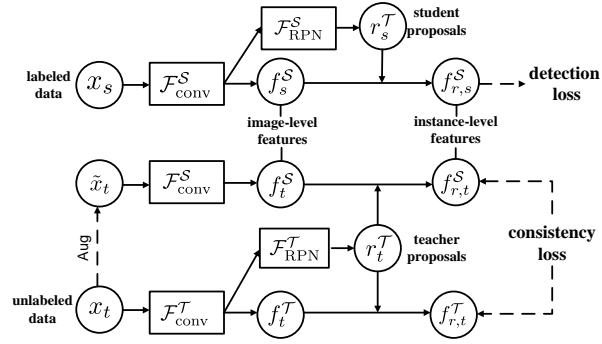}
\caption{The Faster R-CNN with mean teacher framework. For data from the source domain, it is trained in a supervised routine by minimizing the detecting objectives in Faster R-CNN \cite{Faster}. For unlabeled data from the target domain, it is trained by optimizing the consistency regularization between the teacher's and the student's prediction. Teacher detector shares its proposals with the student when training on the unlabeled data.}
\label{FigMTFaster}
\end{figure}

We build the proposed framework MDBank based on Faster R-CNN and mean teacher, as illustrated in Figure \ref{FigMTFaster}.
Notice that we select Faster R-CNN \cite{Faster} as the bare detector for fair comparison with previous methods \cite{MTOR,DA,SWDA}. 

\textbf{Faster R-CNN detector}. Faster R-CNN \cite{Faster} is a two-stage detector, which consists of a feature extractor backbone $\mathcal{F}_{\rm conv}$, a region proposal network (RPN) $\mathcal{F}_{\rm RPN}$ and a region convolutional neural network (RCNN) $\mathcal{F}_{\rm RCNN}$.
For the input data $\bf x$, the image-level features are first calculated by ${\bf f} = \mathcal{F}_{\rm conv}({\bf x})$, and the object proposals are represented by ${\bf r} = \mathcal{F}_{\rm RPN}({\bf f})$.
Then the instance-level features $\bf f_r$ is obtained according to ${\bf r}$, and the bounding boxes \textbf{b} and the category probabilities \textbf{p} can be predicted by $({\bf b,p})=\mathcal{F}_{\rm RCNN}(\bf f_r)$.

\textbf{Mean Teacher in Faster R-CNN}.
Mean teacher \cite{MT} $\mathcal{T}$ is used to provide relatively robust pseudo labels for unlabelled samples, which is established as the same network structure to the student $\mathcal{S}$.
Its parameters $\mathcal{W}^\mathcal{T}$ at $t$-th iteration are updated in a moving average manner: $\mathcal{W}^\mathcal{T}_t={\alpha}\mathcal{W}^\mathcal{T}_{t-1} + (1-\alpha)\mathcal{W}^\mathcal{S}_t$, where $\mathcal{W}^\mathcal{S}_t$ are student's parameters at $t$-th iteration and $\alpha$ is the moving average factor for controlling the update speed of the teacher.
Following \cite{MTOR}, the proposals ${\bf r}^\mathcal{T}$ from the teacher $\mathcal{T}$ are fed into $\mathcal{F}_{\rm RCNN}$ of both the student and teacher detectors, respectively.

For the labeled data $\bf (x_s, y_s)$ from source domain ${\bf D_s}$, the normal supervised routine in \cite{Faster} is applied to train the student detector by minimizing the supervised detection objective $\mathcal{L}_{\rm det}$.
For the unlabeled data $\bf x_t$ from target domain ${\bf D_t}$, the teacher model is used to obtain object proposals ${\bf r}^\mathcal{T}$ and their pseudo labels $({\bf b^\mathcal{T},p^\mathcal{T}})$.
Then the augmented input $\bf \tilde{x}_t$ is fed into the student detector along with ${\bf r}^\mathcal{T}$ to obtain the predictions $({\bf b^\mathcal{S},p^\mathcal{S}})$.
Thus, the consistency regularization of mean teacher could be calculated as following:
\begin{equation}
\label{EqInstanceConsistency}
L_{\rm mt} = L_p({\bf p}_r^\mathcal{T}, {\bf p}_r^\mathcal{S}) + L_b({\bf b}_r^\mathcal{T}, {\bf b}_r^\mathcal{S}),
\end{equation}
where $L_p(\cdot, \cdot)$ and $L_b(\cdot, \cdot)$ are consistency objectives for the classification and bounding box regression between the teacher and the student detector, respectively.

\section{Methodology}
\begin{figure}[t]
\centering
\includegraphics[width=12cm]{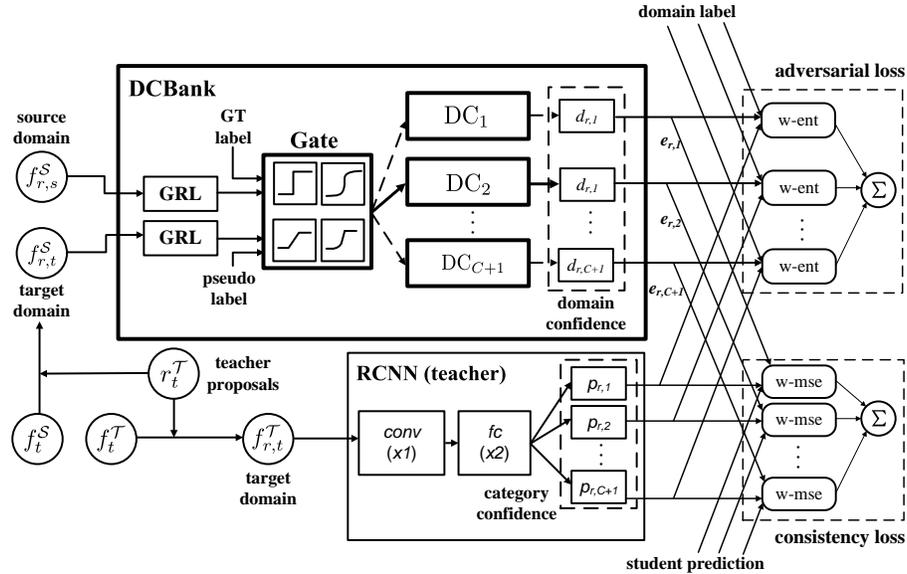}
\caption{The architecture of MDBank for UDA detection tasks. Based on Faster R-CNN detector and mean teacher, the teacher detector shares the proposals with the student to further align the instance-level feature in the same region. DCBank consists of a group of domain classifiers that performs class-level feature alignment in an adversarial learning manner. Crossed adaptive weighting mechanism is applied on instance-level feature $f_r$ between the consistency regularization in mean teacher and adversarial objective in DCBank. The confidence from the second stage of teacher detector $\mathcal{F}_{\rm RCNN}^\mathcal{T}$ is used as a gate function for training the DCBank module while the entropies from the DCBank weight the consistency objective of different categories, respectively. The 'w-ent' and 'w-mse' are the abbreviations of weighted cross-entropy and weighted mean square error, respectively.}
\label{FigOverview}
\end{figure}

In this section, we describe the proposed framework MDBank specifically, shown in Figure \ref{FigOverview}. 

\subsection{Domain Classifier Bank}

The existing adaptive detection methods \cite{DA,SWDA} used a single domain classifier to align instance-level features of different labels, which might prevent the detectors from distinguishing their categories.
To align the features according to their categories, a group of domain classifiers are established and each of them is responsible for aligning features of a specific class.


Formally, we establish the domain classifier bank (DCBank) as $\bf{\mathcal{D}}=\{\mathcal{D}_i\}_{i=1}^{C+1}$, where $C+1$ denotes the number of domain classifiers corresponding to $C$ object categories and the background category.
All classifiers don't share parameters with each other.
%
These domain classifiers are trained to distinguish the instance level features from either the source domain or the target domain.
For the instance level feature ${\bf f}_{r,s}$ from source domain with category label $y_{r,s}$, only the domain classifier $\mathcal{D}_{y_{r,s}}$ in the DCBank is activated to align the region features.
For the instance level feature ${\bf f}_{r,t}$ from the target domain with unknown category, the prediction $p_r^\mathcal{T}$ of the teacher model is referred as the pseudo label to activate the specific domain classifiers to align the region features, as illustrated in Figure \ref{FigOverview}.
For example, the category of maximum predicted confidence $\tilde{y}_{r,t}={\rm argmax}({\bf p_r}^\mathcal{T})$ can be regard as its pseudo label such that only the $\tilde{y}_{r,t}$-th domain classifier $\mathcal{D}_{\tilde{y}_{r,t}}$ is adopted to make the instance level feature alignment.
In fact, a more soft and robust way is to simultaneously select several domain classifiers for domain alignment according to the uncertainty of the teacher model (detailed in Section \ref{SecAdaptiveWeighting}).

According to the label of the instance-level features, domain classifiers can be trained by minimizing the objective $L_{\rm dcbank}$:
\begin{equation}\label{dcbank}
\begin{split}
L_{\bf{\mathcal{D}}}({\bf f_r}) &= \frac{1}{|{\bf r}_K^\mathcal{T}|}\sum_{r{\in}{\bf r}_K^\mathcal{T}}\sum_{i=1}^{C+1}G(y_{r,t}){\rm log}(1 - \mathcal{D}_i({\bf f}_{r,t})) \\
&+ \frac{1}{|{\bf r}_K^\mathcal{S}|}\sum_{r{\in}{\bf r}_K^\mathcal{S}}\sum_{i=1}^{C+1}G(y_{r,s}){\rm log}(\mathcal{D}_i({\bf f}_{r,s})) \\
 &= L_{\bf{\mathcal{D}},s}({\bf f}_{r,s}) + L_{\bf{\mathcal{D}}, t}({\bf f}_{r_t}),
\end{split}
\end{equation}
where $G(\cdot)$ is the activation function to decide which domain classifier is trained.
And the instance-level and class-level feature alignment is achieved by adversarial training:
\begin{equation}\label{EqAdv}
\begin{split}
L_{\rm adv} = \max\limits_{\mathcal{F}_{\rm conv}^\mathcal{S}}&\min\limits_{\bf{\mathcal{D}}}E_{{\bf x} \sim {\bf x_s}}[L_{\bf{\mathcal{D}},s}(\mathcal{F}_{\rm conv}^\mathcal{S}({\bf x})_{\bf r})] \\
&+ E_{{\bf x} \sim {\bf x_t}}[L_{\bf{\mathcal{D}},t}(\mathcal{F}_{\rm conv}^\mathcal{S}({\bf x_t})_{\bf r})].
\end{split}
\end{equation}

That is, the domain classifiers in DCBank try to distinguish the domain label of the instance-level features conditioned by their GT/pesudo labels, while the feature extractors are trained to generate domain-invariant features to cheat those classifiers. Inspired by the Gradient Reversal Layer (GRL) \cite{GRL} where signs of output gradients are flipped, the adversarial loss in Equation \ref{EqAdv} can be easily implemented by adding GRL onto the instance-level features ${\bf f_r}$ before the domain classifier bank module.

Since the DCBank module is deployed to align instance-level features, which are much smaller than the image-level feature, it will add a little storage and computational effort when training. When testing, it can be completely omitted without any addition cost.

\subsection{Crossed Adaptive Weighting} \label{SecAdaptiveWeighting}

Moreover, domain classifiers in DCBank can be trained in a robust manner by incorporating the prediction confidence $\bf{p_{r,t}}$ from the teacher model such that a soft gate function $G(\cdot)$ can be deployed to weight the domain classifiers.
Here, we define two types of the gate function:
\begin{align}
\label{EqGate}
G_1(\bf{p_{r,t}})&={\rm onehot}(\mathop{{\arg\max}}_{1{\leq}c{\leq}C+1}(\bf{p_{r,t}})) \\
G_2(\bf{p_{r,t}})&=\bf{p_{r,t}}^\gamma,
\end{align}
where $G_1(\cdot)$ only activates the domain classifier at the index of maximum prediction score, and $G_2(\cdot)$ activates all the domain classifiers regarding to their confidence score.
Since prediction of the teacher model might be incorrect, DCBank with the soft gate function $G_2$ is believed to be more robust comparing to the hard gate function $G_1$.

In the meanwhile, the prediction of DCBank can reveal the extent of current instance-level feature to be aligned.
The more extent of the feature alignment, the more consistency should be made between the teacher and the student.
Since the domain classifier bank plays a discriminator role by adversarial training, the entropy of the prediction could be used to weight the consistency discrepancy regularization.
Formally, given the output scores ${\bf d_r}=(d_{r,1}, d_{r,2}, \cdots, d_{r, C+1})$ of the DCBanks where $d_{r, i}$ is the output of the $i$-th domain classifier predicted on the $r$-th region-level feature, the entropy of the domain classifier bank can be calculated as
\begin{equation}
\bf{e_r}=-(\bf{p_r}{\log}\bf{p_r}+(1-\bf{p_r}){\log}(1-\bf{p_r})).
\end{equation}
Therefore, the consistency regularization in Equation \ref{EqInstanceConsistency} can be rewritten as
\begin{equation}
\label{EqWeightedInstanceConsistency}
L_{\rm mt} = ||{\bf e_r\odot(p_r^\mathcal{T}-p_r^\mathcal{S})}||_2 + ||{\bf e_r\odot(b_r^\mathcal{T}- b_r^\mathcal{S})}||_2,
\end{equation}
where $\odot$ represents the element-wise product.
For simplicity, we use $l_2$ norm to measure both the classification and bounding box regression discrepancy between the teacher and student detector.
In Figure \ref{FigOverview}, we illustrate the adaptive weighting mechanism between the consistency regularization and the adversarial objective.

\subsection{Overall Objective}
The overall objective consists of three parts: the normal supervised routine of Faster R-CNN $L_{\rm det}$, the weighted instance level consistency regularization from mean teacher $L_{\rm mt}$ and the adversarial objective from DCBank $L_{\rm adv}$:
\begin{equation}
\label{EqObjective}
L_{\rm MDBank} = L_{\rm det} + {\eta}(L_{\rm mt} + {\lambda}L_{\rm adv}),
\end{equation}
in which the parameter $\eta$ controls the weight of MDBank training on the unlabelled data from the target domain, $\lambda$ is the trade-off between output-level alignment from teacher detector and the class-level alignment from MDBank.

\section{Experiments}
We evaluate the proposed method on three public domain shift benchmarks: Sim10k \cite{SIM10K} to CityScape \cite{CityScapes}, PASCAL VOC \cite{PASCAL} to Clipart \cite{PDA} and CityScapes to Foggy CityScapes \cite{Foggy}, which represent three scenarios: synthetic to real scenario, normal to foggy weather scenario and photographic to comic scenario, respectively.

\subsection{Dataset}
SIM10K is a synthetic dataset containing 10,000 training images collected from a synthetic driving game Grand Theft Auto V (GTA5) with bounding box annotation only for cars.
CityScapes is an urban street dataset whose images are captured by a car-mounted camera.
Since these images are annotated in pixel for semantic segmentation task, following \cite{DA,SWDA}, we generate the tightest axis-aligned rectangles as the bounding box labels according to the instance segmentation mask. 
Foggy CityScapes is established based on CityScapes where the images are synthetically rendered with fog according to the depth map, in which each image is rendered in three different levels: $\beta=0.005, 0.01, 0.02$. 
Following \cite{MTOR,SWDA}, the heaviest foggy images ($\beta=0.02$) are used in our experiments.
PASCAL VOC contains images of 20 categories with bounding box annotations. Following the common evaluation protocol, we use both the training and validation split of PASCAL VOC 2007 and 2012 for training, which leads to about 15k images.
Clipart contains 1k comic images sharing the same categories as the PASCAL VOC dataset. Following the setting in \cite{SWDA}, all images are used for unsupervised training and testing.

\subsection{Implement Details}
The backbone network of Faster-RCNN is implemented by ResNet-50 \cite{ResNet} with a Feature Pyramid Network (FPN) \cite{FPN}.
In training stage,
each image is resized as the short size between (960, 1440) pixels while keeping the ratio unchanged.
In particular, images for student model are applied additional augmentation by randomly adjusting the image contrast in scales$(0.5, 1.5)$, the saturation in scales $(0.5,1.5)$ and the brightness in $(-32, 32)$ RGB value. Notice that the above data augmentations keep the objects' categories and bounding boxes unchanged.
The proposed model is trained on 2 GPUs with batch-size=2. We follow the hyper-parameter setting of the base Faster R-CNN to train the images from source domain.
While in target domain, since the labels are unknown, we select the top 512 proposals with highest confidence scores produced by RPN.
Then, those instance-level features are further aligned in the proposed DCBank module.
The moving average weight $\alpha$ for updating the teacher model is set to $0.99$ by default.

\subsection{Declaration for Fair Comparison}
We verify our method by considering the following settings:
\begin{itemize}
  \item The origin faster R-CNN model trained on the source domain without any adaptation is treated our baseline, denoted as Faster.
  \item To evaluate the effectiveness of DCBank, we replace the DCBank with a single instance-level domain classifier without class-level alignment, denoted as ${\rm MT}_{\rm ins}$.
  \item To evaluate the effectiveness of crossed adaptive weighting mechanism, we removes the entropy weighting for consistency regularization and replace confidence weighting for adversarial objective by hard label as the gate function $G_1$ in Equation \ref{EqGate}, denoted as ${\rm MDBank}_{\rm H}$.
  \item The Oracle model directly trained on the target domain in a fully supervised manner, denoted as Oracle.
\end{itemize}
The performance of Source/Target only can be regarded as the lower/upper bound without domain adaptation strategy, which are trained in fully supervised manner.

We also compare MDBank with current state-of-the-arts:
(1) DA \cite{DA} using image-level and instance-level domain classifiers as well as a consistency regularizer, (2) SWDA \cite{SWDA} adopting strong local and weak global feature alignment as well as a context-vector based regularization,
and (3) MTOR \cite{MTOR} incorporating instance-level, inter-graph and intra-graph consistency based on a relationship graph without domain classifiers.
For fairly comparison, we here \textbf{re-implement} DA, SWDA and MTOR with region level consistency 
with the same ResNet-50 with FPN structure.

\subsection{Cross Domain Detection}

\begin{table}[htbp]
\centering
\caption{Experiment results on CityScapes to Foggy CityScapes transfer. The mean average precision (mAP) is evaluated on 8 categories under Foggy CityScapes validation set. The notations G, I, C and AW separately indicate the global-level (or image-level) alignment, instance-level alignment, class-level alignment and adaptive weighting, respectively.}
\begin{tabular}{c|cccc|ccccccccc}
\toprule
method & G & I & C & AW & person & rider & car & truck & bus & train & mcycle & bicycle & mAP \\
\midrule
Faster &  &  &  &  & 30.6 & 41.5 & 40.2 & 6.2 & 38.3 & 48.9 & 7.1 & 13.8 & 28.3\\
DA \cite{DA} & \checkmark & \checkmark &  &  & 39.4 & 48.1 & 48.8 & 31.0 & 42.9 & 54.9 & 7.8 & 18.1 & 36.4 \\
SWDA \cite{SWDA} & \checkmark &  &  &  & 45.8 & 49.2 & 56.2 & 31.1 & 47.0 & 57.5 & 11.2 & 21.9 & 40.0 \\
MTOR \cite{MTOR} &   &   &   &   & 40.4 & 49.7 & 57.6 & 30.1 & 47.9 & 58.6 & 16.9 & 27.1 & 41.0 \\
\midrule
${\rm MT}_{\rm ins}$&   & \checkmark  &   &   & \textbf{46.1} & 48.4 & 54.2 & 30.8 & 45.9 & 56.4 & 20.9 & 25.9 &41.1\\
${\rm MDBank}_{\rm H}$ &   & \checkmark & \checkmark &   & 45.8 & 49.9 & 57.2 & 32.9 & 49.3 & 59.1 & 21.0 & 29.1 & 43.0 \\
\textbf{MDBank} &   & \checkmark & \checkmark & \checkmark & 44.3 & \textbf{50.0} & \textbf{58.4} & \textbf{34.9} & \textbf{48.7} & \textbf{59.1} & \textbf{26.1} & \textbf{28.7} & \textbf{43.8} \\
\midrule
Oracle &   &   &   &   & 46.4 & 54.0 & 65.7 & 41.3 & 54.6 & 64.8 & 34.4 & 30.5 & 48.9 \\
\bottomrule
\end{tabular}
\label{TbC2F}
\end{table}

\subsubsection{Normal to foggy weather images}
The comparison results for normal weather to foggy weather (CityScapes to Foggy CityScape) domain are summarized in Table \ref{TbC2F}, which illustrates the AP of 8 common objects in urban street scenario and their mean AP (mAP).
Our MDBank achieves the best 43.8\% mAP with the margin 2.8\% comparing to the second runner MTOR \cite{MTOR}.
Since the foggy images in target domain are rendered directly from source domain, the domain gap between them is smaller than other scenarios. Naturally, the teacher model can provide more convincible predicting results on unlabeled images.
Thus the proposed MDBank with hard pseudo label ${\rm MDBank}_{\rm H}$ shows similar performance as using soft label 43.0\% v.s. 43.8\% mAP.
We also note that MTOR and ${\rm MT}_{\rm ins}$ achieve similar performance, which indicates that applying a single domain classifier to simply align instance-level feature has limited effects for adaptive object detection.
In summary, with the help of DCBank, our MDBank surpasses MTOR and ${\rm MT}_{\rm ins}$ by around 2.7\% mAP. In Figure \ref{FigResultVis}, we illustrate the detection result under Foggy Cityscapes.

\begin{table}[htbp]
\centering
\caption{Experiment results on SIM10K to CityScapes transfer. The average precision (AP) is evaluated on the car category. The notation G, I, C, AW is following Table \ref{TbC2F}.}
\begin{tabular}{c|cccc|c}
\toprule
method & G & I & C & AW & car AP on target \\
\midrule
Faster &  &  &  &  & 34.9 \\
DA \cite{DA} & \checkmark & \checkmark &   &   & 43.1 \\
SWDA \cite{SWDA} & \checkmark &   &   &   & 47.8\\
MTOR \cite{MTOR} &  &   &   &   & 54.9\\
\midrule
${\rm MT}_{\rm ins}$ &   & \checkmark &   &   & 55.8 \\
${\rm MDBank}_{\rm H}$ &  & \checkmark & \checkmark &   & \textbf{56.3} \\
\textbf{MDBank} &   & \checkmark & \checkmark & \checkmark & \textbf{56.3} \\
\midrule
Oracle &   &   &   &   & 65.9 \\
\bottomrule
\end{tabular}
\label{TbM2C}
\end{table}

\subsubsection{Synthetic to real images}
We first evaluate the performance of MDBank on the synthetic (SIM10K) to real (CityScapes) domain. Table \ref{TbM2C} shows the performance under CityScapes validation set of average precision rate (AP) on car category.
Since there are only one category (car) to be detected, the MDBank performs closed to the ${\rm MT}_{\rm ins}$ and ${\rm MDBank}_{\rm H}$, where DCBank only consists of two domain classifiers.
Own to the class-level feature alignment by DCBank, our MDBank still outperforms MTOR by 1.4\%, which achieves the best 56.3\% of AP. 

\subsubsection{Photographic to comic images}
In this experiment, we analyze our method by evaluating on images from real (PASCAL VOC) to artistic (Clipart) domain.
The evaluation results are shown in Table \ref{TbP2C} with the AP of 20 common objects and their mAP. Though the source and target domain are much dissimilar, as illustrated in Figure \ref{FigResultVis}, our MDBank method also achieves best performance of 45.4\% mAP, which improves about 3\% comparing to current best result achieved by MTOR.
Note that ${\rm MT}_{\rm ins}$ even performs worse than the MTOR, which reveals that the simple instance-level alignment regardless of categories information might have side effect as the number of category growing.
MDBank surpassing ${\rm MDBank}_{\rm H}$ by 1.5\% verifies the effectiveness of the crossed adaptive weighting strategy. Since the all images are used for unsupervised adaptive training \cite{SWDA}, the performance of the oracle is meaningless that it is trained on the validation set in a fully supervised manner.

\begin{table}[t]
\centering
\caption{Experiment results on dissimilar domain transfer from PASCAL VOC to Clipart datasets. The mean average precision (mAP) is evaluated on 20 categories on all 1k images in Clipart. The mAP of oracle is only for referenced since all the images from target domain are used for training following the setting in \cite{SWDA}. The table is transposed for fully visualization.}
\begin{tabular}{c|cccc|ccc|c}
\toprule
method & Faster\cite{Faster}\quad & DA\cite{DA} \quad& SWDA\cite{SWDA}\quad & MTOR\cite{MTOR}\quad & ${\rm MT}_{\rm ins}$ \quad & ${\rm MDBank}_H$\quad & MDBank\quad & Oracle \\
\bottomrule
areo & 18.6 & 12.7 & 37.0 & 39.9 & 33.5 & 40.2 & \textbf{42.5} & 62.5\\
bycle & 34.6 & 40.3 & 67.2 & 74.5 & 67.1 & \textbf{76.0} & 70.4 & 77.8\\
bird & 17.1 & 21.4 & 30.6 & 22.8 & 24.8 & 26.3 & \textbf{37.8} & 77.0\\
boat & 11.2 & 16.2 & 28.3 & 39.4 & \textbf{42.7} & 39.4 & 38.7 & 60.5\\
bottle & 23.8 & 35.7 & 44.6 & 45.8 & \textbf{51.3} & 38.5 & 48.8 & 61.1\\
bus & 44.3 & 29.3 & 65.0 & 58.8 & 55.9 & \textbf{69.1} & 58.0 & 86.6\\
car & 23.4 & 29.6 & 41.9 & 56.3 & 56.2 & \textbf{59.1} & 57.7 & 81.8\\
cat & 11.3 & 0.6 & 13.1 & 13.0 & 13.2 & 9.7 & \textbf{19.5} & 67.1\\
chair & 35.5 & 29.6 & 52.2 & 56.9 & 58.0 & \textbf{58.2} & 52.9 & 77.9\\
cow & 5.2 & 39.1 & 42.7 & 18.6 & 31.2 & \textbf{36.9} & 30.6 & 76.9\\
table & 22.8 & 18.4 & 22.3 & 36.7 & 34.9 & 31.6 & \textbf{36.5} & 77.4\\
dog & 6.0 & 10.9 & 7.2 & 5.8 & 13.9 & 10.8 & \textbf{15.3} & 75.1\\
horse & 21.2 & 19.7 & 26.9 & 29.4 & 30.0 & 32.2 & \textbf{38.0} & 74.8\\
bike & 40.8 & 61.5 & 76.3 & 79.0 & 58.9 & 80.2 & \textbf{85.1} & 87.2\\
person & 29.0 & 54.1 & 53.1 & 58.7 & 63.9 & 65.8 & \textbf{66.3} & 85.5\\
plant & 35.4 & 41.2 & 55.0 & \textbf{64.5} & 57.1 & 56.9 & 57.1 & 73.2\\
sheep & 0.4 & 16.9 & 11.2 & 11.4 & 8.9 & \textbf{18.3} & 17.3 & 76.6\\
sofa & 18.2 & 16.5 & 24.1 & 19.9 & \textbf{25.6} & 24.8 & 17.2 & 69.0\\
train & 24.9 & 16.3 & 48.9 & 56.3 & 45.1 & 41.9 & \textbf{58.5} & 70.4\\
tv & 22.8 & 33.7 & 51.2 & \textbf{62.0} & 58.6 & 61.2 & 59.4 & 81.3\\
\midrule
mAP & 22.3 & 27.2 & 40.0 & 42.5 & 41.5 & 43.9 & \textbf{45.4} & 75.0\\
\bottomrule
\end{tabular}
\label{TbP2C}
\end{table}

\begin{figure}[htbp]
\centering
\subfigure[]{
\begin{minipage}{0.45\textwidth}
\centering
\includegraphics[height=3.5cm]{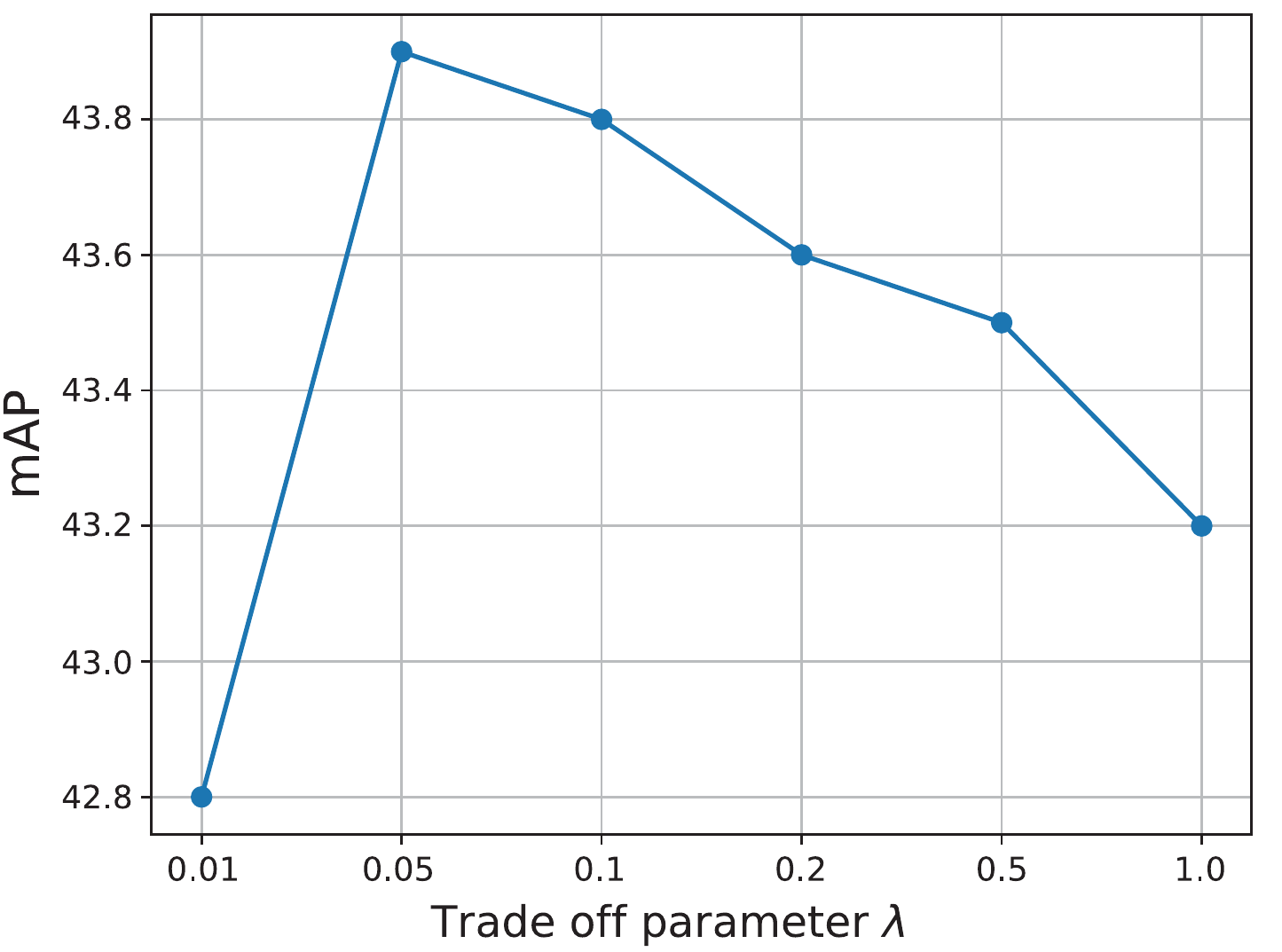}
\end{minipage}
}%
\subfigure[]{
\begin{minipage}{0.45\textwidth}
\centering
\includegraphics[height=3.5cm]{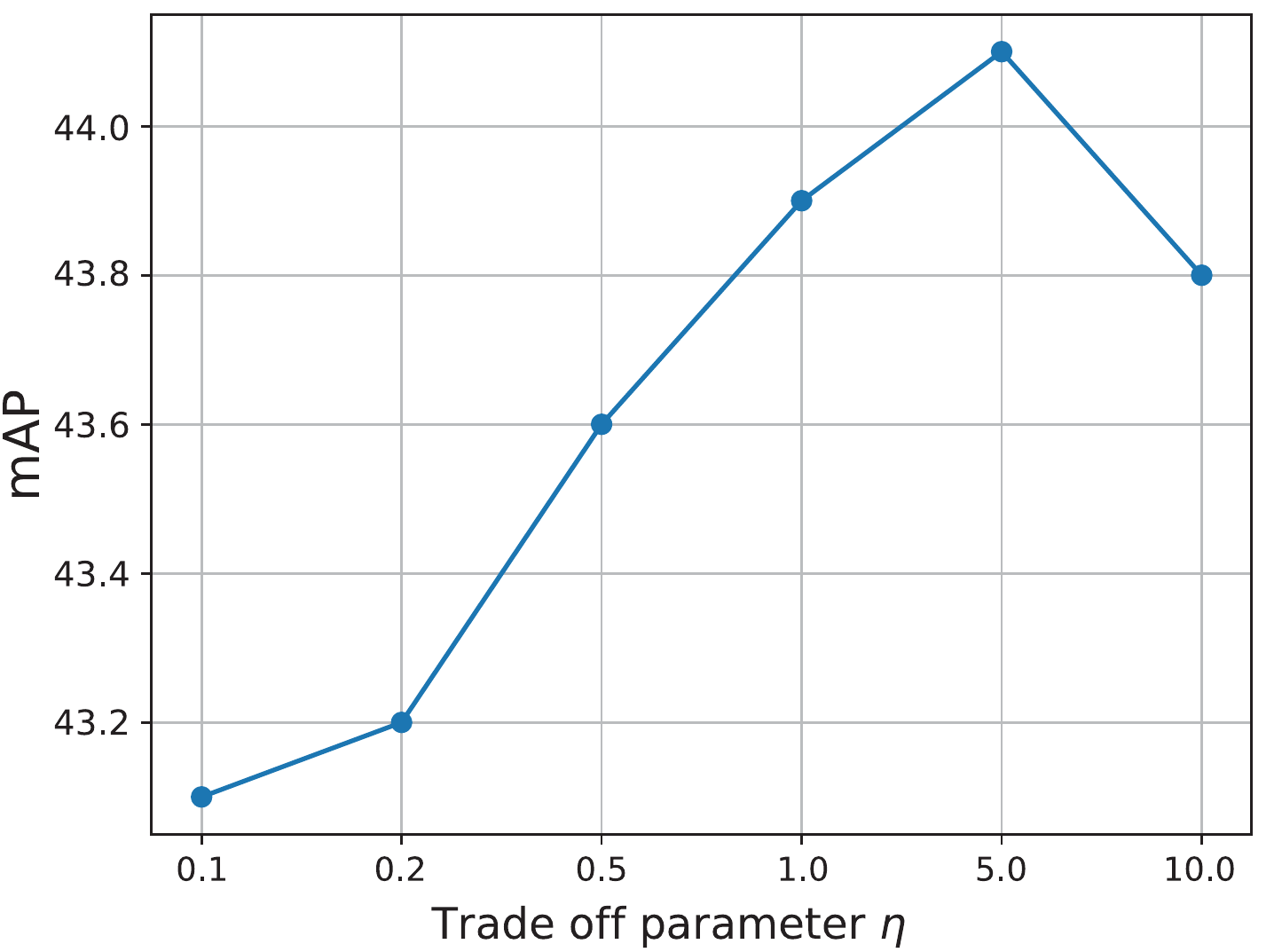}
\end{minipage}
}
\caption{The effect of the trade off parameters $\lambda$ and $\eta$ under CityScapes to Foggy CityScapes transfer.}
\label{FigAblHyper}
\end{figure}

\subsubsection{Analysis of hyper-parameter $\lambda$ and $\eta$}
$\lambda$ controls the trade-off of the student detector learned from the source and target domain, while $\eta$ weights the objective of the DCBank module.
Figure \ref{FigAblHyper} shows the mAP under different value of hyper-parameters $\lambda$ and $\eta$, respectively.
We find that MDBank is relatively robust to the trade off parameter $\lambda$ in a wide range.
MDBank can achieve the best 44.1\% mAP when $\lambda=0.1$ and $\eta=5.0$.

\subsubsection{Visualization on the target domain}
\begin{figure}[htbp]
\centering
\includegraphics[width=12.3cm]{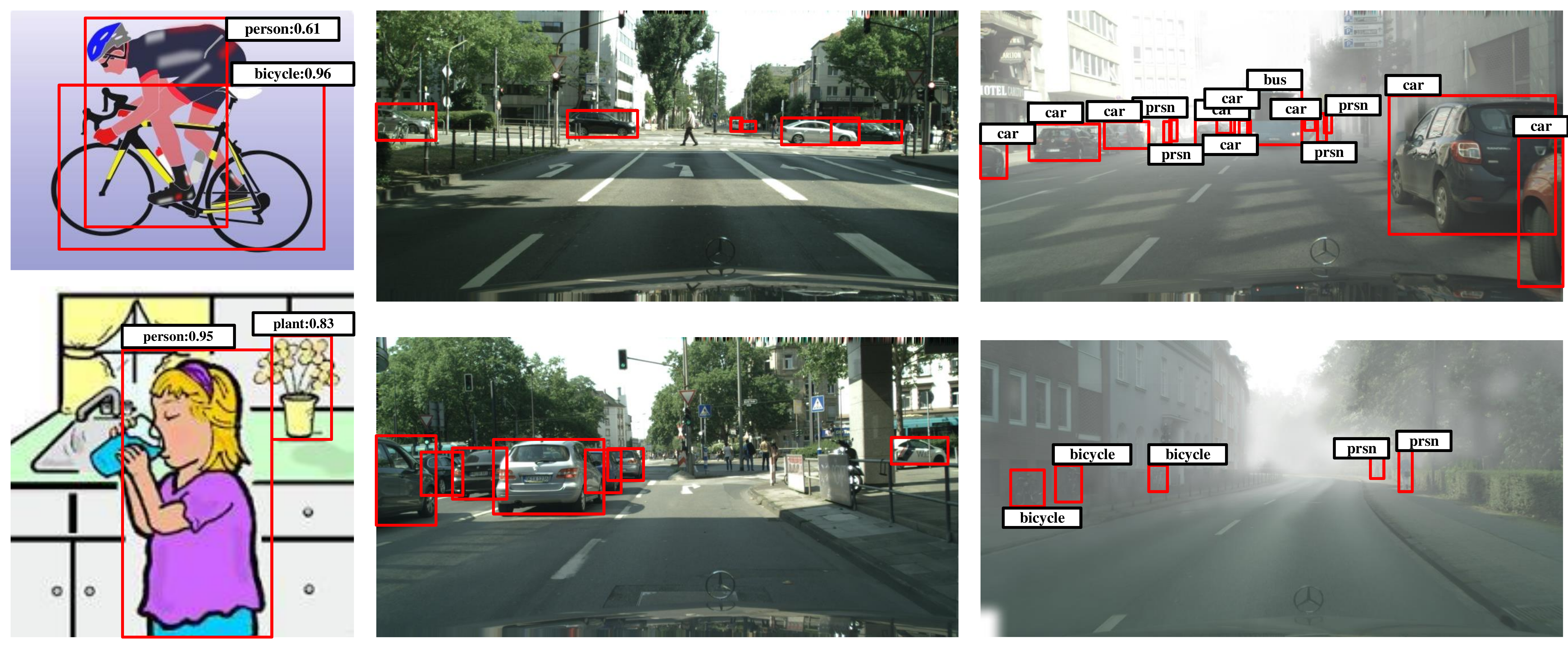}
\caption{The examples of detection results from the proposed MDBank on the target domain. From left to right is PASCAL VOC to Clipart, SIM10K to CityScapes and CityScapes to Foggy CityScapes transfer. For clarity, the confidence score is omitted on images from CityScapes and Foggy Cityscapes. Notice that 'car' is the only object to be detected in SIM10K to CityScapes transfer.}
\label{FigResultVis}
\end{figure}

\begin{figure}[htbp]
\centering
\subfigure[{\scriptsize DA/domain}]{
\begin{minipage}{0.23\textwidth}
\centering
\includegraphics[height=2.8cm]{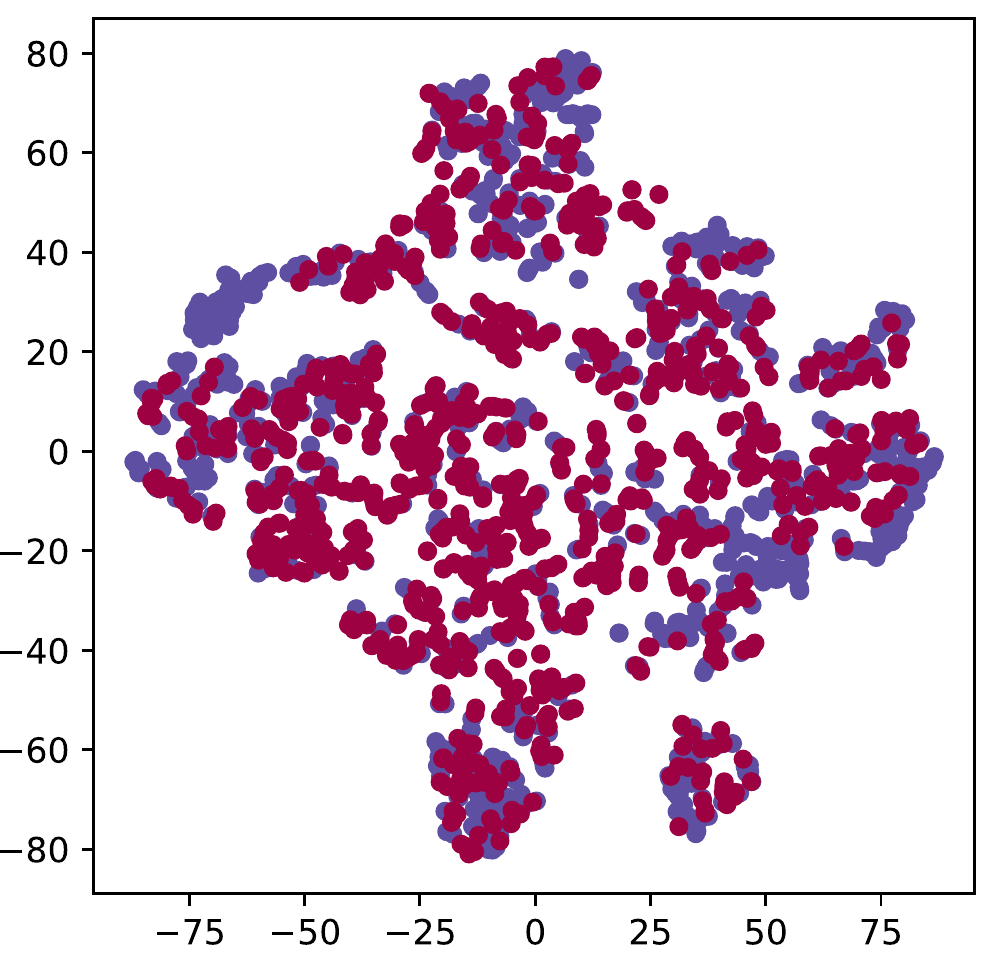}
\end{minipage}
}%
\subfigure[{\scriptsize DA/category}]{
\begin{minipage}{0.23\textwidth}
\centering
\includegraphics[height=2.8cm]{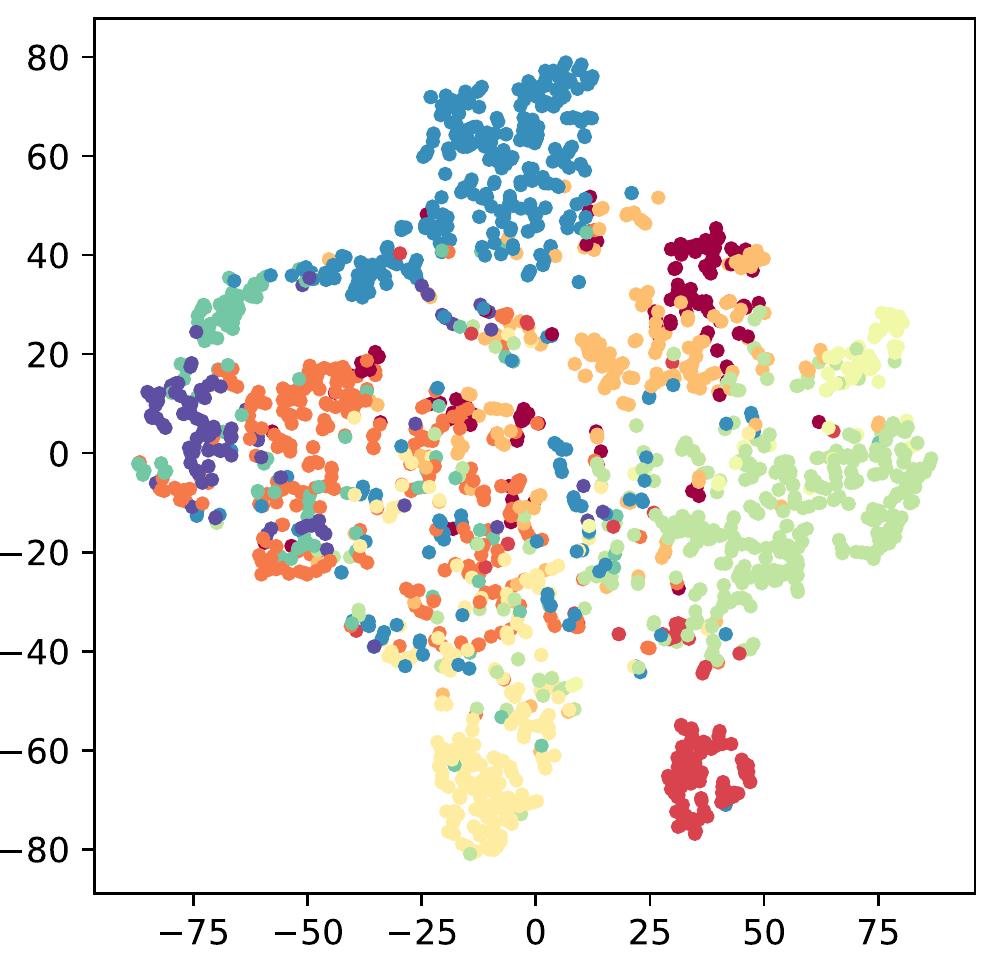}
\end{minipage}
}%
\subfigure[{\scriptsize MDBank/domain}]{
\begin{minipage}{0.23\textwidth}
\centering
\includegraphics[height=2.8cm]{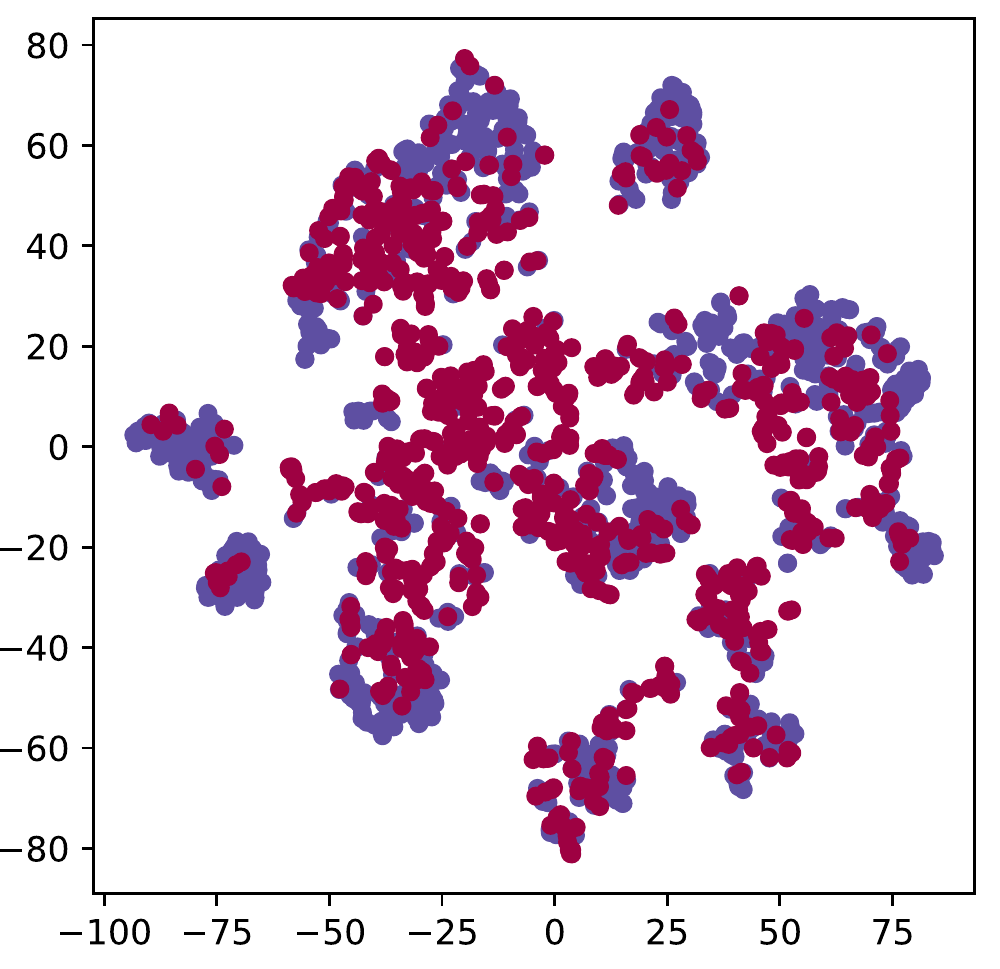}
\end{minipage}
}%
\subfigure[{\scriptsize MDBank/category}]{
\begin{minipage}{0.23\textwidth}
\centering
\includegraphics[height=2.8cm]{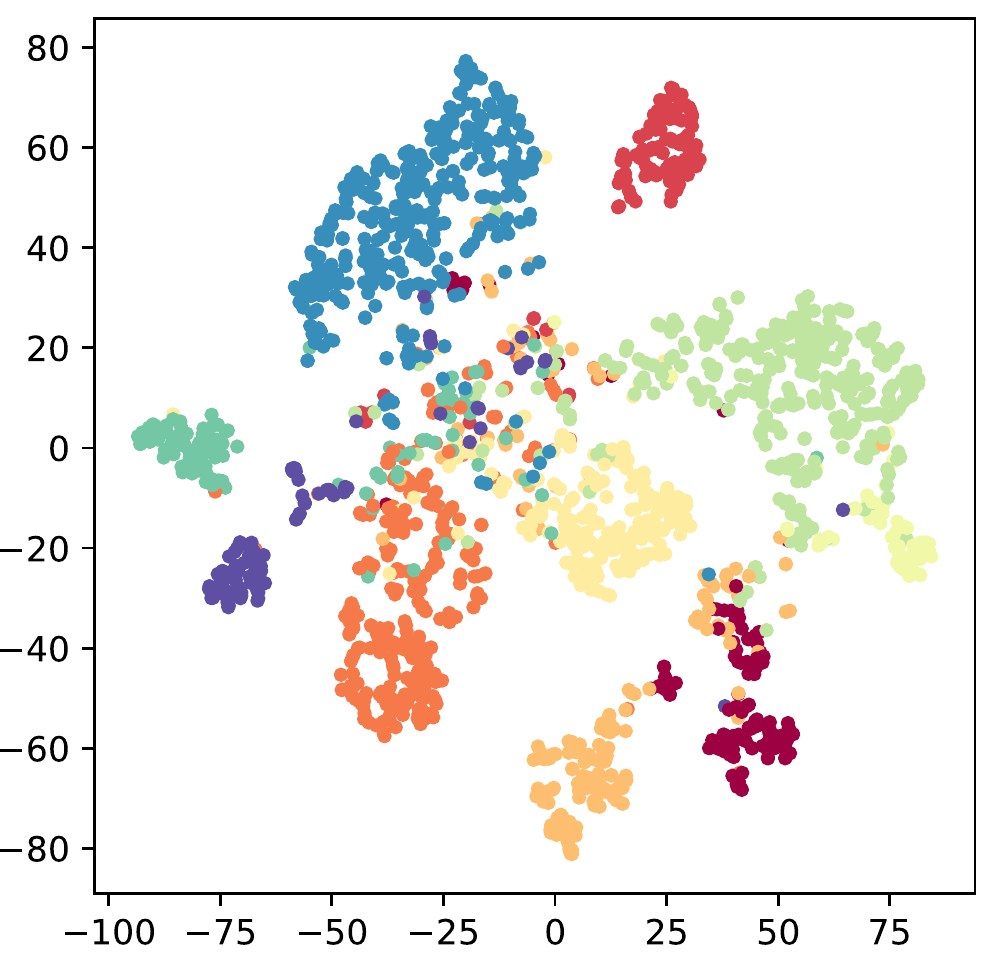}
\end{minipage}
}%
\caption{Evidences of instance-level feature distribution on PASCAL VOC to Clipart transfer.
Features belonging to the first 10 classes in Table \ref{TbC2F} are illustrated for better visualization. Here we compare the proposed MDBank and DA\cite{DA}.
(a) and (c) show the instance-level feature distribution with red/blue color representing instances from source/target domain.
(b) and (d) show their category labels with different colors.
Although the instance-level features are well-aligned both in MDBank and DA by referring to (a) and (c), the MDBank achieves more accurate class-level alignment by referring to (b) and (d).
}
\label{FigTSNEP2C}
\end{figure}

In Figure \ref{FigResultVis}, we illustrate examples of detection results by the proposed MDBank on the target domain. The detection results show that the MDBank is relative robust to the similar and dissimilar domain transfer. In Figure \ref{FigTSNEP2C}, we show the instance-level feature distribution of the proposed MDBank and DA \cite{DA}.
Notice that DA applies instance-level domain alignment without considering their categories. Though both MDBank and DA achieve similar alignment referring to their domain labels as in Figure \ref{FigTSNEP2C} (a) and (c), our MDBank shows more distinguishable boundaries among instances of different categories as illustrated in \ref{FigTSNEP2C} (d).
Since there are much more instance categories than the previous transfer tasks, MDBank outperforms DA with a large gap by applying the class-level feature alignment. 

\section{Conclusions}
In this paper, we present an MDBank framework including a mean teacher with a domain classifier bank for domain adaptation object detection problem in unsupervised manner. 
Our key contribution is the domain classifier bank module that respectively aligns the instance-level features according to their category labels. To align the unlabelled data from target domain, a mean teacher paradigm is incorporated to provide robust pseudo labels while applying instance-level prediction consistency between the teacher and student detector. 
Besides, a crossed weighting mechanism is then proposed to adaptively connect the DCBank and mean teacher to boost their performance. 
Experiment shows that our MDBank achieves the new state-of-the-arts on CityScapes, Foggy CityScapes, SIM10K, PASCAL VOC and Clipart Datasets for unsupervised domain adaptation detection.

%
%
\bibliographystyle{splncs04}
\bibliography{egbib}
\end{document}